# Hate Me Not: Detecting Hate Inducing Memes in Code Switched Languages

*Completed Research Full Papers*

**Kshitij Rajput***
Netaji Subhas University of Technology
rajput.kshitij97@gmail.com

**Raghav Kapoor***
Netaji Subhas University of Technology
raghavk.co@nsit.net.in

**Kaushal Kumar Rai***
Netaji Subhas University of Technology
kaushalk.co@nsit.net.in

**Preeti Kaur**
Netaji Subhas University of Technology
preeti.kaur@nsut.ac.in

## Abstract

The rise in the number of social media users has led to an increase in the hateful content posted online. In countries like India, where multiple languages are spoken, these abhorrent posts are from an unusual blend of code-switched languages. This hate speech is depicted with the help of images to form "Memes" which create a long-lasting impact on the human mind. In this paper, we take up the task of hate and offense detection from multimodal data, i.e. images (Memes) that contain text in code-switched languages. We firstly present a novel triply annotated Indian political Memes (IPM) dataset, which comprises memes from various Indian political events that have taken place post-independence and are classified into three distinct categories. We also propose a binary-channelled CNN cum LSTM based model to process the images using the CNN model and text using the LSTM model to get state-of-the-art results for this task.

## Introduction

Over time people have extensively started using multimodal means to express their sentiments as they believe it is a much stronger way to depict the accurate sense as to how they feel about a particular person or a situation. According to an article[1], Facebook status updates with images get 2.3 times more engagement than status updates without images. It is much easier for a human to process the visual information and about 90% of all information that we perceive is visual[2].

A large proportion of these posted images give rise to the problem of hate speech. With the growing popularity of social media, hate-inducing and violent content are growing exponentially. In many cases, this hate is expressed in Memes by the means of *sarcasm*. Users often club these images with text to convey their angst. Hate speech detection in such imageries is an intricate task as it involves the analysis of images as well as of the text within it.

Hate speech detection in Indian languages is a complex problem due to its rich linguistic diversity. The world of social media has also given rise to code-switched languages. In India, a particular pair of code-mixed language, *Hinglish*, is most popularly used. Hinglish language consists of non-fixed grammar, irregular semantics and spellings. In Hinglish language, the words are written in the Roman script instead of the *Devnagari* script. However, the meaning of the words is in Hindi. The same word can be written in many ways. For example, in the sentence *"ye bahut swaad hai"*, the word *"swaad"*, which means tasty, can be written as *"swad", "swaad", "svaad", "svad"* which all mean the same thing. Also, the word-to-word translation of the sentence would be "This very tasty is" which is grammatically incorrect in English. Hence, as we see, the task gains even more complexity when exhibited in code-switched languages. The vast number of social media users, the rise in hate speech, ambiguities in the semantics and grammar of Hinglish, a labyrinth of image analysis is what demonstrates the magnitude of the problem.

---

[1] https://bit.ly/3fCjGYr

[2] http://bit.ly/2Ht3ubm

[*] These authors contributed equally to the paper.

In this paper, we present a deep learning solution to solve all obstacles to classify the images (Memes) on social media in one of the three categories: *Hate Inducing*, *Satirical* and *Benign* or *Non-Offensive*. We extract the text from the images and process the text and the image independently, and finally combine the results to get the final category in which the complete image with text can be classified. We use LSTM based model for the text classification. For text classification different word embedding models have been tried including Glove (Pennington et al. 2014), FastText (Bojanowski et al. 2017), Word2Vec (Godin et al. 2015), Bert (Devlin et al. 2018). The images have been analyzed with the help of a CNN-based model. A doubly annotated dataset consisting of Indian Political memes which are classified in three categories has been created and released along with the model.

The significant contributions in our work are as follows:

(i) Creation of IPM (Indian Political Memes) dataset.

(ii) Demonstrating how independent analysis of text and images and then recombining the results gives significantly better results than considering the image alone.

(iii) Creation of a deep learning-based classifier model which will outperform all the other baseline models on the IPM dataset.

## Background and Related Work

The initial task for hate speech detection led to the development of a prototype system *Smokey* (Spertus et al. 1997) for detecting email flames (angry or offensive emails) using 47 elements feature set which captured the syntax and semantics of the sentences present in the dataset. Further, libSVM (Yin et al. 2009) is used as the classifier model with local features, sentiment features and contextual features for detecting harassment on Web 2.0. An SVM-based model trained on a corpus of 1,655,131 user comments on Yahoo buzz, combined with valence analysis for detecting personal insults on social news websites was also put forward (Sood et al. 2012). Research (Gamback et al. 2017) proposed a Convolution Neural Network (CNN) for classification on Twitter text data into four categories: sexism, racism, both (racism and sexism) and non-hate-speech using the following features: character 4-grams, word2vec vectors to capture semantic information, randomly generated word vectors, and word vectors combined with character n-grams. A dataset of about 25K tweets labeled as hate-inducing, offensive or benign was released wherein a logistic regression model with L2 regularization was used for classifying the tweets in one of the three categories (Davidson et al. 2017).

However, most of the research on hate speech detection in the past was restricted to English text only. A study (Del et al. 2017) conducted the task of hate speech detection on the Italian language using the following features: (i) morpho-syntactical features, (ii) sentiment polarity and (iii) word embedding lexicons was shown by. However, the task of hate speech detection on code-switched data has its own intricacies of having to deal with non-fixed spellings, grammar, and semantics for this language.

Since our work consists of hate speech detection of memes with text in Hinglish language, so we look at some past work for hate speech detection focusing on Hinglish language. The task of hate speech detection on Hindi-English code-switched data using a Random Forest (RF) classifier and a Support Vector Machine (SVM) classifier was performed previously by Bohra et al. (2018) using the following features: (i) character n-grams, (ii) word n-grams, (iii) punctuations, (iv) negation words and (v) lexicons. A ternary trans CNN model (Mathur et al. 2018) using transfer learning was also developed for hate speech detection on Hindi-English code-switched dataset. Further work proposed (Kapoor et al. 2018) on hate speech detection on Hindi-English code switched HEOT dataset using LSTM based model with transfer learning which takes Glove embeddings and Word2Vec embeddings as features.

In the recent past there has also been some research on detecting sarcasm from code switched languages using SentiWordNet (Gupta et al., 2021). Furthermore, Jain et al. (2020) uses a BiLSTM-CNN model to achieve an accuracy of 92.71% on code switched (Hinglish) tweets for sarcasm detection. However, the above text-based models exhibit limited accuracy while dealing with the intricacies of images.

Wei et al. (2006) focused on analyzing sentiment out of images where they proposed a novel mechanism of finding the emotion out of an image by finding an orthogonal three-dimensional factor space of an image and then passing it through an SVM classifier. Siersdorfer et al. (2010) analyzed the relation between the sentiment of images expressed in metadata and their visual content in the social photo-sharing environment Flickr.

## Dataset and Evaluation

### Dataset Acquisition

We constructed an Indian Political Memes (IPM) Dataset for this experiment which consisted of memes that are shared on a day-to-day basis on the internet and are politically motivated. We created this dataset using the *"google_images_download"*[3] module which is an open-source python tool available online, to scrape several images using the keywords as the name of some famous politicians, social activists, journalists, and big political events that have taken place post-independence in India. For each keyword 100 images were downloaded using the module, resulting in a corpus of images containing 5000 images. Out of this corpus of images, 1500 memes were randomly sampled and were asked to be annotated by three annotators into the following categories:

(i)     Hate Inducing
(ii)    Satirical
(iii)   Non-offensive.

Some of the images that were blurred and consisted of no text were removed from the dataset resulting in a final dataset of 1218 memes. The memes were annotated as hate inducing if and only if the meme satisfied one or more of the following conditions: (i) meme consisted of a sexist or racial barb to malign a minority, (ii) meme had object stereotyping or (iii) meme consists of a hateful hashtag such as "*#HinduSc\*m*". The annotators were specifically asked not to consider a meme as hate-inducing due to the presence of a particular word, however offensive that word might be.

| Label | IPM Dataset |
|---|---|
| Non-Offensive | 339 |
| Hate-Inducing | 427 |
| Satirical | 452 |
| Total | 1218 |

**Table 1: Class Distribution in Indian Political Memes (IPM) dataset**

Once all the annotators had labelled each image in the dataset in one of the three categories, all the conflicts were resolved and finally, the label that was in majority was chosen as the final label for the meme. The distribution of the memes into the various classes is shown in Table 1. This dataset was then channelized into a pipeline that extracts the text from the images which is further explained in Methodology section.

---

3 http://bit.ly/2Jv55Qe

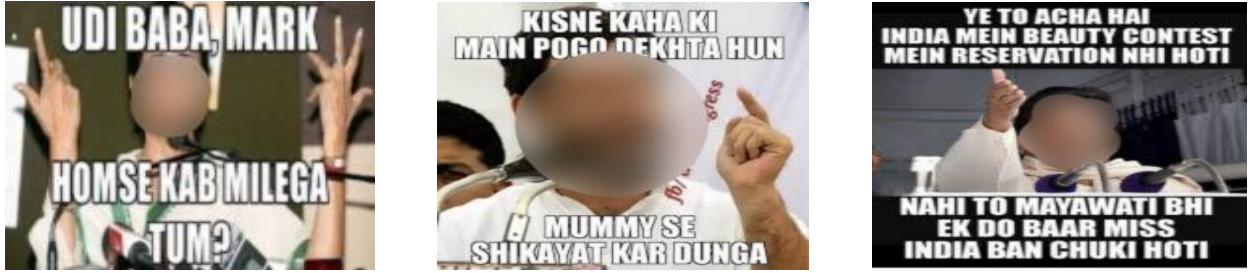

**Figure 1: Examples of a) Non-Offensive b) Satirical and c) Hate-Inducing Memes in IPM**

After final annotation, it was found that there were 427 hate-inducing memes, 452 satirical memes, 339 non-offensive memes out of the total 1218 memes dataset. Examples of non-offensive, satirical and hate-inducing memes are shown in Figure 1. This dataset was then channelized into a pipeline that extracts the text from the images. Table 2 shows the text extracted from the image, their English translation, and the label assigned to them. We have calculated several the value for multiple evaluation metric for our dataset.

| Figure | Hinglish Text Extracted | English Translation | Label |
|---|---|---|---|
| Figure 1 (a) | Udi baba, Mark, homse kab milega | Hey Mark, when will you meet us | Non Offensive |
| Figure 1 (b) | Kisne kaha ki main pogo dekhta hun. Mummy se shikayat karunga | Who said that I watch pogo. I will complain to mother | Satirical |
| Figure 1 (c) | Ye to acha hai India mein beauty contest mein reservation nhi hoti | It is good that there in no reservation in beauty contests in India (Derogatory remark on personal appearance) | Hate Inducing |

**Table 2: Example of Hinglish text extraction from the memes with their respective English translations**

The Cohen's Kappa score between the three annotators for our dataset is shown in Table 3. The highest kappa score of 0.87 between annotator A2 and annotator A3 is shown. The multilingual index which is used to calculate inequality in the diffusion of languages in a corpus containing two or more languages was found to be 0.684 for our dataset, indicating good code-switching. The Fleiss's Kappa score for our dataset was 0.782 indicating decent agreement between the three annotators.

| | A1 | A2 | A3 |
|---|---|---|---|
| A1 | _ | 0.81 | 0.77 |
| A2 | 0.81 | _ | 0.87 |
| A3 | 0.77 | 0.87 | _ |

**Table 3. Cohen's Kappa for the three annotators A1, A2, A3**

# Methodology

## Preprocessing Images

The image needs to be pre-processed to extract text out of the meme efficiently. Though the OCR performs some inbuilt pre-processing, we perform the following steps for processing the image ourselves:

(i) Rescaling: Some of the images need to be rescaled to a larger size to enlarge the text written in a smaller font in the meme to make the text recognizable to the OCR.

(ii) Gaussian blurring: The blurring effect is used to reduce noise from the image. Gaussian blurring is performed by convolving the image with a gaussian kernel.

(iii) Deskewing: Some of the memes have text written at some skewed angle. Deskewing helps to rotate the image such that the text written in the image is mostly horizontal.

(iv) Gaussian adaptive thresholding: Thresholding \cite{stauffer1999adaptive} converts the text into a black and white format so that it is easily recognizable to OCR. In adaptive thresholding, the gaussian mean of the surrounding area determines the threshold value for the pixel of the image.

## Extraction of text from images

After performing the above pre-processing on the images, we pass it through an open-source OCR reader *ocr.space*[4] for extracting the text out of the memes. Table 2 depicts the samples after text extraction from the image examples given above along with their English translations.

## Preprocessing Text

The tweets obtained from data sources were sent through a pipeline with the objective to convert them into semantic feature vectors.

(i) Initially, the hashtags (For example: *#indianpolitics*), URLs, user mentions (denoted by '@') and numbers were removed from the text since they do not convey any relevant information about the sentiments of the text. Also, using the NLTK library, the stop words were eliminated.

(ii) The emoticons (For example: ":)", "XD") were replaced by their textual description about the true emotions they depict.

(iii) Many of the comments which are in *Devnagari* (Hindi) script were converted to Roman (English) script. This was done using a python library called *indic-transliterate*[5]

(iv) The Hinglish text now obtained are converted to their respective English translation using an *Xlit-Crowd Conversion Dictionary*[6].

(v) This is followed by the use of various word embedding representations such as FastText Bojanowski et al. 2017, Twitter word2vec (Godin et al. 2015), Glove (Pennington et al. 2014) and Bert (Devlin et al. 2018) embeddings for building the first layer of the LSTM side of the model, the word-embedding layer. Different embeddings models are used to obtain the word vector representations of the preprocessed tweets. The embedding models are used one by one to figure out the best set of word embeddings.

## *The Model*

Taking inspiration from previous works described in Rajput et al (2020), we propose a novel method which involves a binary channeled CNN-cum-LSTM model that takes the text in the form of word vector representation and image as its input and finally concatenates the two channels to produce the result. The model architecture is depicted in Figure 2.

---

[4] http://bit.ly/3ouxnQ9

[5] http://bit.ly/2JtSc95

[6] http://bit.ly/2WVCeY8

**The CNN Channel**

The CNN channel processes the image and tries to extract certain features of the image that would help to classify the meme into one of the three categories i.e., hate inducing, satirical and non-offensive. The pre-processed images form the input to the first layer of the CNN channel which is a convolution2D layer with filter size 64, kernel size as (5, 5) and activation function of *Relu* (Nair et al. 2010). This is followed by a max_pooling layer of size (5, 5). Next, we employ another convolution2D layer, this time of size 32, kernel size (3, 3), activation function as *Relu*, and a max_pooling layer of pool size (3, 3). This is succeeded by a flatten layer that converts the 3-dimensional feature map to 1 dimension. We also use a dropout layer of size 0.4 to prevent overfitting of data. This is followed by a dense layer of size 32. The CNN channel tries to utilize the image features to decide in which category the meme is to be classified.

**The LSTM Channel**

The first layer of the LSTM channel is the embedding layer which takes the word vector representation of the extracted caption from the Meme image. These embeddings help to learn distributed representations of captions. After experimentation, we kept the size of embeddings fixed to 100. Different embedding models unravel the different aspects of the language. For example, the dependency parser focuses on the similarity between the two terms. On the other hand, statistics of bag of word (BoW) embeddings emphasize on word associations. Our work on bag of words was inspired by the usage of it proposed in the research by Rai et al (2019). The embedding layer is followed by a dropout layer of size 0.2 to prevent overfitting of data. The next is the LSTM layer of size 64 with a dropout of 0.4. The LSTM layer is followed by two dense layers of sizes 64 and 32. This part of the model serves as a processing model for the textual content.

**Combination of Channel**

The two parallel channels of the model which process the text and images separately are finally recombined to a single channel to obtain the results. The concatenation is followed by the presence of two dense layers of sizes 32 and 3. The last dense layer of size 3 uses *softmax* as the activation function. We use L2 regularization and *Adam* optimizer (Kingma et al. 2014) for preventing overfitting. The loss function used in the last layer was categorical cross-entropy. The output obtained after passing through the dense layers is one of the three classes, i.e., Hate Inducing, Satirical and Non-offensive.

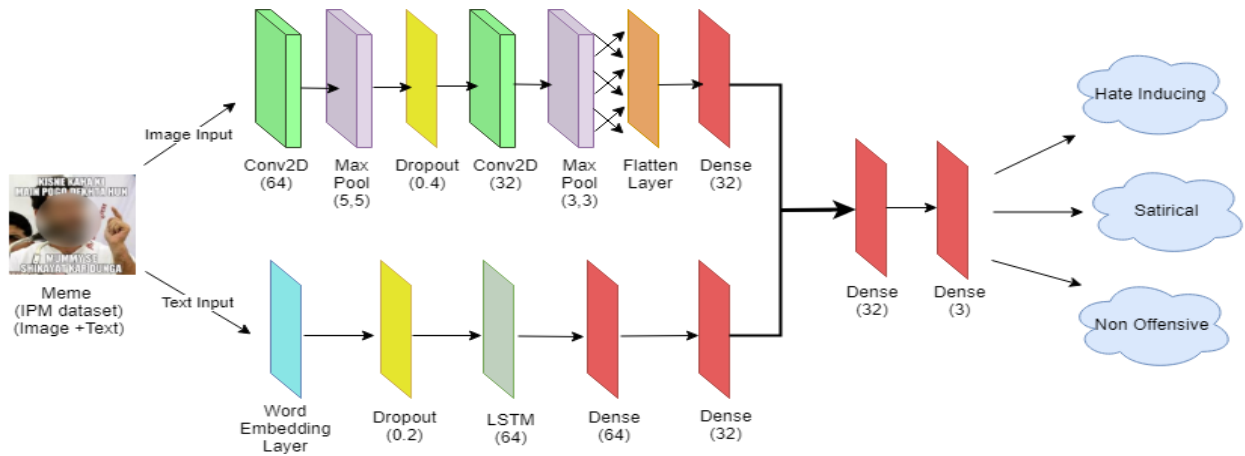

**Figure 2: Proposed model architecture**

# Experimentation and Results

In this section, we analyze the results of various models on the IPM dataset. As a baseline, we first conduct experiments using supervised machine learning models, namely SVM and random forest classifier which will define the baseline results. We then use LSTM and CNN models followed by the analysis of our proposed model.

### Baseline

The baseline model was created using a Support Vector Machine (SVM) and a Random Forest (RF) classifier. These two classifier models were trained using k-fold cross-validation with 10 splits. For the SVM classifier, we choose kernel value as 'poly' with the default value of degree = 3 as hyperparameters. All other hyperparameters for the SVM classifier are used at their default values. For the RF classifier after fine-tuning the model, the hyperparameters chosen were n_estimators as 600, max_depth as 12 and max_features as *log2*. We choose the following features from the images to be input in the baseline classifying models:

(i) GLCM features: Gray-Level Co-occurrence Matrix helps to determine the texture of the image which is useful for determining the emotional expression in an image. We use GLCM for determining the contrast, correlation, energy, and homogeneity of an image (Haralick et al. 1992) which will be used as features for our model.

(ii) Colorfulness feature: Color is one of the important ways to convey a message through an image. Colorfulness is calculated using Earth Mover's Distance (EMD) between the histogram of an image and the histogram having a uniform color distribution (Datta et al. 2006).

(iii) Tamura features: Tamura features also help for determining the texture of an image (Tamura et al. 1978). We use coarseness and directionality as Tamura features for input to the classifying model.

(iv) Human face feature: Human faces are important for drawing attention to a meme. The number of human faces and the size of human faces are used as features of our classifying model. (Viola et al. 2004).

The results using these baseline models and the features mentioned are shown in Table 4. We use precision, recall and F1-score as the metrics for determining the baseline results. It was seen that SVM performs marginally better than the Random Forest classifier when using GLCM features. The SVM classifier also gave comparable results when using the human face features as the input. This forms our baseline results for Hinglish offensive memes classification on IPM Dataset.

| Feature | GLCM | | Colorfulness | | Tamura | | Human Face | |
|---|---|---|---|---|---|---|---|---|
| Classifier | SVM | RF | SVM | RF | SVM | RF | SVM | RF |
| Precision | 0.622 | 0.584 | 0.542 | 0.475 | 0.562 | 0.512 | 0.608 | 0.545 |
| Recall | 0.651 | 0.597 | 0.522 | 0.457 | 0.592 | 0.538 | 0.619 | 0.595 |
| F1-Score | 0.634 | 0.575 | 0.573 | 0.514 | 0.583 | 0.546 | 0.658 | 0.603 |

**Table 4: Baseline Results on IPM Dataset using SVM and Random Forest**

### Deep Learning Model

We also compare our model to some of the deep learning models. We use two deep learning models for validating the accuracy of our own model. The first model we experimented with, is a CNN-based model which is popular for image classification. The CNN model was trained using a k-fold cross validation with 10 splits. The first layer of the CNN model is a Convolution2D layer with filter size 64 and kernel size = (5, 5). The activation function used is *Relu* (Nair et al. 2010) and a max_pooling layer of pool size = (5, 5) is employed. This is followed by another Convolution2D layer with a filter size of 32 and kernel size of (3, 3), succeeded by a max-pooling layer of pool size = (3, 3). We also employ a dense layer of size 64 and a dropout layer of 0.4 to prevent overfitting. The hyperparameters are chosen with the help of grid search which helps to select those hyperparameters that produce the most optimal results.

The second model experimented with is an LSTM based model which consists of an embedding layer, followed by an LSTM layer and two dense layers of size 64 and 32. Adam optimizer and L2 regularization are used to prevent overfitting of data. The dropout layer of size 0.4 is also added to prevent overfitting of data. The LSTM based model is also trained using k-fold cross-validation with 10 splits. The loss function used for both the CNN-based model and the LSTM based model is categorical cross-entropy.

| Result | CNN Model | LSTM Model |
|---|---|---|
| Precision | 0.632 | 0.581 |
| Recall | 0.674 | 0.534 |
| F1-Score | 0.618 | 0.604 |

**Table 5: Result of Deep Learning Model on IPM Dataset**

We compare the results of the two deep learning models in Table 5 using precision, recall and F1-score as the metrics. The CNN-based model produces much better results than the LSTM based model, while the results are marginally better than the SVM baseline model described above. This is due to the fact that hate speech can be conveyed in the form of text as well as images. Analyzing the image solely therefore does not produce great results for this classification task. Hence, we see that the features extracted manually and fed to SVM or RF classifier give comparable results to that of CNN or LSTM based models. As proposed in our model, we analyze both images as well as the text extracted to produce the results.

## *Our Model*

Now we compare the results of the baseline (SVM and RF) model and the deep learning models (CNN and LSTM) to our proposed model. Our model tries to incorporate the best features of both the CNN and LSTM deep learning models by considering images as well as the extracted text from the images for the classification task. We propose a binary channel model in which the LSTM channel processes the text written inside the images and the CNN channel processes the image itself. The two channels are combined to produce the final results. We conduct the experiments on our model using different flavors of word embeddings i.e. (i) Twitter Word2vec (Godin et al. 2015) (ii) Glove (Pennington et al. 2014) (iii) Fastext (Bojanowski 2017) and (iv) Bert (Devlin et al. 2018) embeddings as well as the combination of the above embeddings. Many different sizes of embeddings were tested. Finally, the size of the embeddings was chosen to be 100. Our model is also trained using k-fold cross-validation with 10 splits to maintain consistency in all the experiments.

The results of our model using different types of word embeddings on the IPM dataset are shown in Table 6. The results are compiled using precision, recall and F1-score as metrics of evaluation. Our model outperforms the baseline (SVM and RF models) and also the deep learning (CNN and LSTM) models, hence establishing itself as the state of the art for the task of Offensive memes classification in Hinglish language. As seen from Table 6, the best results obtained using a single embedding model were with the Glove embeddings. A recall score of 0.816 was recorded with Glove embeddings. Also, experiments were conducted using the combination of word embeddings where (Glove + Fastext) is seen to produce the best results. Here, a precision of 0.798 was obtained which demonstrates that the model outperforms all the other models on the IPM dataset for the task of offensive memes classification in code switched language (Hinglish).

| Features | Precision | Recall | F1-Score |
|---|---|---|---|
| Glove (Gl) | 0.762 | 0.816 | 0.792 |
| Twitter Word2vec (Tw) | 0.741 | 0.766 | 0.781 |
| FastText (Ft) | 0.721 | 0.694 | 0.773 |
| Bert (Bt) | 0.758 | 0.804 | 0.784 |
| (Gl) + (Tw) | 0.727 | 0.751 | 0.722 |
| (Gl) + (Ft) | 0.798 | 0.779 | 0.764 |
| (Tw) + (Bt) | 0.748 | 0.740 | 0.702 |
| (Gl) + (Bt) | 0.779 | 0.790 | 0.725 |
| (Bt) + (Ft) | 0.760 | 0.723 | 0.746 |

**Table 6: Result of our model with different word embeddings**

### *Error Analysis*

We analyze the possible reasons due to which our model gives error in its judgment.

(i) OCR error: We have used *ocr.space* for extracting the text out of the memes. The memes in which the text is written in very small font, or the text written is slightly blurred, the OCR fails to recognize that text with 100% accuracy. Also, in many cases, it is hard for the OCR reader to extract text which is written vertically or diagonally.

(ii) Unconventional words (code switched): A little work is done in dealing with uncommon Hinglish words which may arise due to spelling variations, grammatical errors or mixing of some regional languages by the creators of the memes. For example, the spelling variation resulting from a difference in the pronunciation of the words can create a new set of words that are not present in the dictionary itself.

(iii) Disguised hate: Some memes are designed so that they might seem to be satirical to the annotators but might be inducing hate towards an individual in a disguised fashion. Such memes would not be correctly classified by our model. For example, *"Abbe oh, ma\*\*\*sa jane vale"* which translates to *"Hey, religious school going person"*

## Conclusion

In this paper, we introduced a novel dataset, i.e the IPM (Indian Political Memes) which consists of images (Memes) classified in three categories - Benign, Satirical and Hate Inducing. Also, we proposed a pipeline to detect offense and hate from images that contain text in code-switched languages. We developed a multi-channel CNN-LSTM model, which processes the images and text individually and combines the analysis from both channels to give the final classification result. Several different words embedding models are tried to attain the most optimum results. The images are passed through the CNN channel of the model. We compare the results of our model to other deep learning-based models and some supervised machine learning models, namely SVM and Random Forest classifier after extracting features from the images. The results suggest that our model outperforms all the other models producing state-of-the-art results for Hinglish language Memes classification on the IPM dataset. We also release the code, the dataset made, and the model proposed in our work. We believe this method would be useful for hate speech detection for images in code-switched languages.